# Application of transfer learning to sign language recognition using an inflated 3D deep convolutional neural network

**Roman Toengi**
The Open University
rt3726@ou.ac.uk

## Abstract

Sign language is the primary language for people with a hearing loss. Sign language recognition (SLR) is the automatic recognition of sign language, which represents a challenging problem for computers, though some progress has been made recently using deep learning. Huge amounts of data are generally required to train deep learning models. However, corresponding datasets are missing for the majority of sign languages. Transfer learning is a technique to utilize a related task with an abundance of data available to help solve a target task lacking sufficient data. Transfer learning has been applied highly successfully in computer vision and natural language processing. However, much less research has been conducted in the field of SLR. This paper investigates how effectively transfer learning can be applied to isolated SLR using an inflated 3D convolutional neural network as the deep learning architecture. Transfer learning is implemented by pre-training a network on the American Sign Language dataset MS-ASL and subsequently fine-tuning it separately on three different sizes of the German Sign Language dataset SIGNUM. The results of the experiments give clear empirical evidence that transfer learning can be effectively applied to isolated SLR. The accuracy performances of the networks applying transfer learning increased substantially by up to 21% as compared to the baseline models that were not pre-trained on the MS-ASL dataset.

## 1 Introduction

### 1.1 Sign language recognition

Sign language (SL) is expressed entirely visually through the three principal channels of hand movements, facial expression and body postures with the potential issue of occlusion where one channel hides another one. Sign language recognition (SLR) is the task of recognizing human sign gestures. Computer-based SLR has been researched for many years, although every research has its limitations. There are somewhere between 138 and 300 sign languages worldwide (Brooks, 2018), but most research focuses on one particular sign language employing tailored datasets.

While historically sign languages had not been considered to be true languages at all, modern linguistic research (Stokoe, 1960) has recognized that sign languages are complete natural languages with their own grammar and lexicon like spoken languages. Sign languages are not universal, they differ substantially from each other, and they are mutually unintelligible (Woll, 2013). Most sign languages make use of both hands, where one hand is the dominant one, and the other hand is the passive one. This results in 1) one-handed signs formed by the dominant hand only, 2) two-handed symmetrical signs where the handshapes and movements of the two hands are symmetrical, and 3) two-handed non-symmetrical signs performed by moving the dominant hand and the passive hand serving as a base (Lapiak, 2013). SLR systems have to take this into account to utilize the rich features of sign language fully. Nevertheless, most research neglects the aspect of SL linguistics completely (Er-Rady et al., 2017).

There exist various data acquisition methods of two major categories: vision-based and sensor-based (Cheok et al., 2019). Common vision-based methods are 1) single (i.e. monocular) colour camera, 2) stereo camera to provide depth information, 3) active techniques using the projection of structured light, and 4) intrusive techniques using some form of body markers. Sensor-based methods require the use of special instruments that range from 1) inertial measurement units determine the position, orientation and acceleration of body parts, 2) electromyography measures human muscle's electrical pulses to detect movements, 3) Wi-Fi and radar detect in-air signal strength changes, and 4) other technologies including ultrasonic, mechanical, electromagnetics and haptic sensors.

SLR can be implemented at the word or sentence level. Isolated SLR processes segmentations of single signs, which is contrasted by continuous SLR handling whole SL sentences (Al-Shamayleh et al., 2018). Further, SLR can be categorized as either signer-dependent or signer-independent, the latter meaning that the signers used during training are not allowed to appear in the dataset used to evaluate the system (Ibrahim et al., 2018). Desirably, such systems also perform well for unknown signers.

In conclusion, SLR is not yet a solved problem in general, facing the challenges of visual complexity, lack of adequate SL training datasets, and underutilization of SL linguistics. As a result, current SLR systems work well in a laboratory setting on limited synthetic datasets but perform poorly in a realistic environment.

## 1.2 Transfer learning

Representation learning extracts meaningful information from the input data so that subsequent prediction models can perform their tasks more easily. Deep learning, in which a hierarchy of concepts is learned through the composition of multiple nonlinear transformations, are highly suitable to learn meaningful representations, as depth is considered a key aspect for successful representation learning. Deep structures offer the following advantages: 1) the promotion of the reuse of features, and 2) leading to more abstract features at higher levels (Bengio et al., 2013). Representation learning is advantageous to transfer learning tasks, as they learn representations that capture the underlying explanatory factors of the data, which may be relevant for each particular transfer learning task.

Neural networks usually require lots of data to achieve satisfactory results. However, for many problems, the necessary amount of data is either not available or cannot be easily generated. In such cases, transfer learning might be employed, which utilizes a different but related task with an abundance of data available to help solve a target task lacking enough data. In this way, a model trained for a related task can be repurposed to a different task, and thereby transferring the knowledge it has already acquired from another domain. For example, deep neural networks trained on images exhibit an interesting phenomenon: irrespective of the concrete image dataset, each trained network learns similar features at lower layers of the network, which are more general and become more specific deeper in the network (Yosinski et al., 2014). Since the more general features are invariant to specific tasks within a domain, they can be utilized to transfer knowledge across those tasks.

## 2 Related work

Different technological strands have developed over time to build SLR systems.

Hidden Markov models have been widely used on sequence-based problems, such as speech recognition. Starner and Pentland (1995) employed a hidden Markov model for the development of a proof-of-concept system to recognize American Sign Language from videos. Von Agris and Kraiss (2007) developed a video-based automatic SLR using a hidden Markov model to classify German Sign Language. As part of their work, they created a new sign language corpus, claiming to meet the requirements for signer-independent SLR, called SIGNUM.

Fang et al. (2017) developed a translator for American Sign Language at both the word and sentence level, called DeepASL. They used an infrared light-based sensing device for data acquisition with the built-in capability to extract the joints of the forearms and hands of a signer, which is followed by a bi-directional deep recurrent neural network with LSTM nodes used for classification.

Xue et al. (2019) proposed a vision-based SLR system, which was trained on the American Sign Language Lexicon Video Dataset (Athitsos et al., 2008). Their system uses OpenPose (Cao et al., 2019) to detect body joints as a pre-processing step, which is followed by a random forest classifier.

A lot of research has been done on inventing effective and efficient deep architectures for image-related tasks. Instead of repeating the work for video-related tasks, Carreira and Zisserman (2017) proposed the inflated 3D ConvNet architecture, called I3D, which extends the existing 2D ConvNet Inception-v1 network trained on the ImageNet dataset (Szegedy et al., 2015) by a temporal dimension. Highly interestingly, they not only expanded the structure of Inception-v1 but also reused a scaled version of its trained parameters. They trained I3D on the human action classification dataset Kinetics (Kay et al., 2017) improving significantly upon the state-of-the-art performances on the existing benchmarks for human action classification.

Joze and Koller (2019) tackled the problem of American Sign Language recognition proposing the first large-scale dataset for American Sign Language, called MS-ASL. They identified the I3D network as a suitable architecture for sign language recognition but recognized that pre-training the I3D model on the Kinetics dataset is not beneficial for the task of sign language recognition.

Our work is mainly influenced by Carreira and Zisserman (2017) and Joze and Koller (2019). Whereas Carreira and Zisserman (2017) applied 2D ConvNet inflation to the Inception-v1 network, we inflated the improved successor network Inception-v3 (Szegedy et al., 2016). As pre-training on a human action classification dataset, such as Kinetics, is not advantageous for sign language recognition (Joze and Koller, 2019), our inflated 3D ConvNet was only bootstrapped with the ImageNet trained parameters. To implement transfer learning we used MS-ASL as the dataset for the source task and SIGNUM as the dataset for the target task.

## 3 Methodology

This work evaluates how effectively transfer learning can be applied to isolated SLR. We adopted the single colour camera as the capturing method, which results in a sequence of colour images as the modality for the network's input data. This data representation was selected as it is non-intrusive, allowing for widespread usage without the need for special equipment. The TensorFlow machine learning framework was used to implement the neural networks, which were trained on a single Nvidia Titan RTX graphics card utilizing the CUDA platform. The full source code is publicly available in the accompanying GitHub repository (Toengi, 2020), which is distributed under the MIT license.

### 3.1 Overall method

This paper investigates the extent that knowledge learned from one SLR task is transferable to a different SLR task. Specifically, the learned features of a model recognizing American Sign Language (ASL) were utilized by a model recognizing German Sign Language (GSL). Transfer learning was carried out in the following two steps:

1. A model was pre-trained on the MS-ASL dataset to implement the SLR task of American Sign Language.

2. The pre-trained model was then fine-tuned on the SIGNUM dataset to implement the SLR task of German Sign Language.

To evaluate the effectiveness of applying transfer learning, the predictive performance of the fine-tuned model was compared to a model trained only on the SIGNUM target dataset, referred to as the baseline model. This resulted in the following two different types of models:

- Baseline model that did not apply transfer learning being trained only on the SIGNUM target dataset.

- Fine-tuned model that applied transfer learning being pre-trained on the MS-ASL source dataset and subsequently fine-tuned on the SIGNUM target dataset.

### 3.2 Common architecture to both models

The basic network architecture common to both types of models was based on the Inception-v3 2D ConvNet with an input image size of 224x224 pixels. This network was inflated by another dimension to capture the temporal information in the data. We adopted an initial temporal depth of 20, which was collapsed steadily to 6 before the last pooling layer.

This inflated network was bootstrapped with the ImageNet trained parameters of Inception-v3 in order to provide low-level spatial features, such as edge detectors, to both the baseline and fine-tuned models. Otherwise, transfer learning as applied in this work would have been trivially effective because the model pre-trained on the MS-ASL source dataset could have better learned this basic spatial capability. However, this work focuses on evaluating the effect of transfer learning at a higher level improving the learning of the task-specific features of sign language recognition.

### 3.3 Datasets

MS-ASL was used as the source dataset, which consists of a total of 25,513 links to YouTube videos of isolated ASL signs. However, at the time of downloading, some videos were no longer available since the associated YouTube account had been terminated meanwhile, for example. Thus, the source dataset used in this work contained 22,063 examples in total. This dataset is already provided as separate training, validation and test datasets. To make use of the entire MS-ASL data for pre-training, the three dataset splits were combined into one data collection. As a result, there was no validation dataset used during training. To prevent the model from overfitting, training was stopped when the model's accuracy reached 95%.

The SIGNUM target dataset contains recordings of 25 different signers. The first signer is denoted as the reference signer who performed each sign three times. However, to not introduce any bias with respect to the reference signer, only his first performances were used. This gave a total of 11,250 recordings of isolated GSL signs (each of the 25 signers performed each of the 450 signs once).

From each example of both datasets, 20 frames were uniformly sampled at random to match the temporal depth of the network. Each frame was also cropped to the central square around the signer, resized to the network's input image size of 224x224 pixels, and finally, the RGB values were scaled to the interval [-1, 1]. Such a sequence of 20 frames denoted one dataset example fed into the network. As an example, the following figure depicts the 20 frames (arranged row-wise) for each of the ASL gesture and the GSL gesture both for the word *dance*:

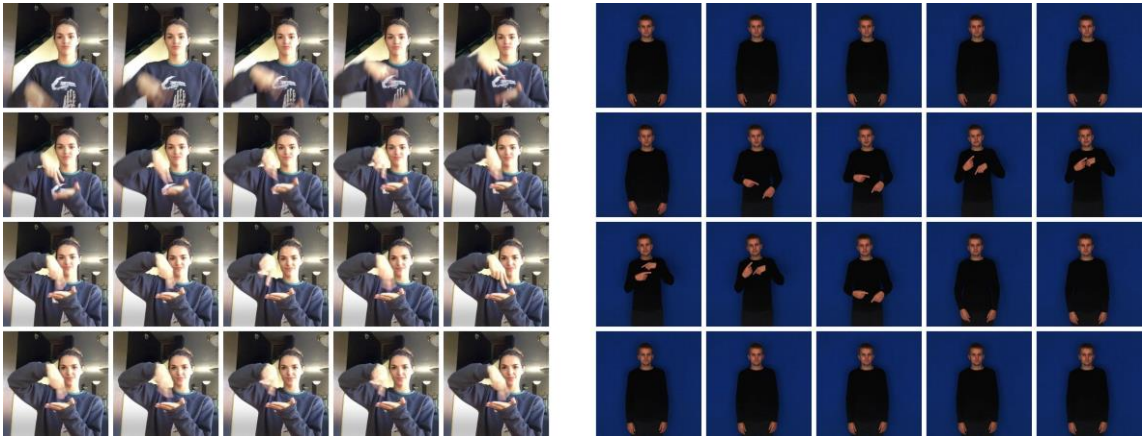

Figure 1: Input frames of the ASL sign (left) and the GSL sign (right) both for the word dance. As can be seen, the MS-ASL dataset is comprised of realistic photographic material as compared to the SIGNUM dataset, which had been recorded in a very controlled setting. Note how the gesturing of the word dance differs between American Sign Language (left) and German Sign Language (right).

To assess the impact of the size of the target dataset on transfer learning, experiments with three different sizes of SIGNUM training datasets were carried out, referred to as the large, medium and small training dataset, respectively. Further, the initial SIGNUM dataset was split in a signer-independent manner. As for the large dataset setting, the recordings of the first 16 signers went into the training dataset, the recordings of the next 4 signers went into the validation dataset, and the recordings of the remaining 5 signers went into the test dataset. As for the medium and small dataset settings, the training datasets were halved and quartered, respectively, while the validation and test datasets were the same as with the large dataset setting. The following table summarizes the signer-independent splitting for the three dataset sizes:

Table 1: Signer-independent SIGNUM dataset splits. The counts given are the numbers of examples that the corresponding dataset contained.

| Signer | 1 2 3 4 5 6 7 8 9 10 11 12 13 14 15 16 | 17 18 19 20 | 21 22 23 24 25 |
|---|---|---|---|
| Dataset | Training | Validation | Test |

| Dataset size | | | |
|---|---|---|---|
| Large | 7,200 | 1,800 | 2,250 |
| Medium | 3,600 | 1,800 | 2,250 |
| Small | 1,800 | 1,800 | 2,250 |

### 3.4 Training procedure

As outlined, one type of model was trained solely on the SIGNUM target dataset, referred to as the baseline model. The number of classes of the output layer was set to 450 to correspond to the number of sign classes in the SIGNUM dataset. This model was trained on each of the large, medium and small SIGNUM training dataset, resulting in three corresponding trained baseline models.

The other type of model applying transfer learning was first pre-trained on the MS-ASL source dataset by setting the number of classes of the output layer to 1,000 to coincide with the number of sign classes in the MS-ASL dataset. Subsequently, the number of classes of the output layer of the pre-trained model was adapted from 1,000 to 450 to meet the requirement of the SIGNUM dataset. This model was then fine-tuned separately on each of the large, medium and small SIGNUM training dataset as had been carried out with the baseline models in precisely the same manner. The outcome was three fine-tuned models having applied transfer learning.

The inflated Inception-v3 network bootstrapped with the ImageNet trained parameters was the network architecture for both the baseline and fine-tuned models. All the neural networks were trained to minimize the categorical cross-entropy loss using standard SGD with momentum set to 0.9 with a batch size of 32. Further, each model was trained for 40 epochs so that the resulting learning curves could be better compared. However, a model's state was saved and used later for evaluation at the epoch at which the model achieved the lowest loss value on the SIGNUM validation dataset. This application of early stopping prevented the networks from overfitting.

### 3.5 Evaluation of the effectiveness of applying transfer learning

The hypothesis was that the model that applied transfer learning is superior to the baseline model with respect to the accuracy performance on the target test dataset. As outlined, three models that involved transfer learning had been trained:

- Model pre-trained on MS-ASL and fine-tuned on *large* SIGNUM training dataset
- Model pre-trained on MS-ASL and fine-tuned on *medium* SIGNUM training dataset
- Model pre-trained on MS-ASL and fine-tuned on *small* SIGNUM training dataset

To have a reference for comparison, the following three baseline models had been trained that did not apply transfer learning:

- Model trained only on *large* SIGNUM training dataset
- Model trained only on *medium* SIGNUM training dataset
- Model trained only on *small* SIGNUM training dataset

Each of the six models above was evaluated on the SIGNUM test dataset resulting in six corresponding test accuracy scores. The accuracy score of each model applying transfer learning was compared to the accuracy score of the baseline model that had been trained on the same size of the SIGNUM training dataset. This allowed for three evaluations of the effectiveness of transfer learning with respect to the impact the size of the target training dataset had.

# 4 Experiments and results

## 4.1 Comparison of the training processes between the baseline and fine-tuned models

Thanks to bootstrapping the initial model architecture with the ImageNet trained parameters, the training of the models converged in a small number of epochs. Transfer learning exhibited the following three beneficial effects on the training processes: 1) better initial predictive performance, 2) faster convergence, and 3) better final predictive performance. The comparisons of the training processes with respect to the accuracy metric between the baseline and fine-tuned models for each size of the training datasets are given below:

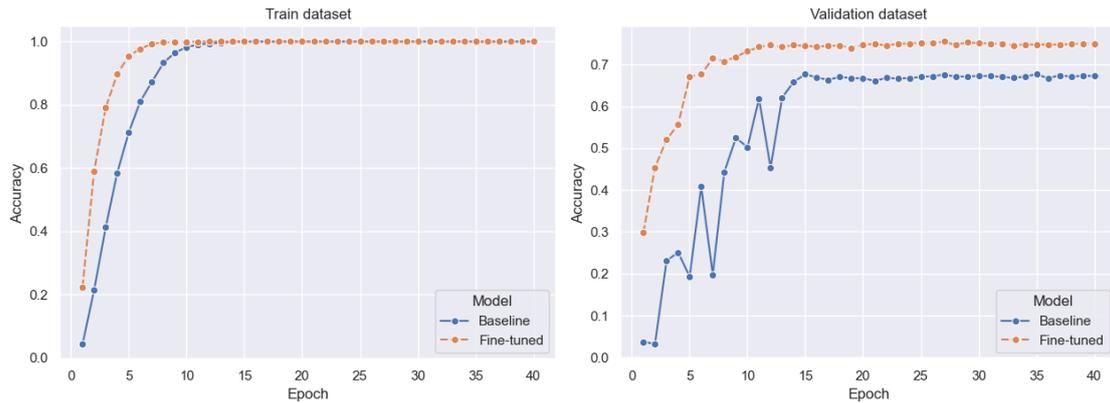

Figure 2: Comparison of the training processes with respect to the large target dataset. The fine-tuned model's accuracy value converged at about `0.75` on the validation dataset, whereas the corresponding value for the baseline model was only about `0.67`.

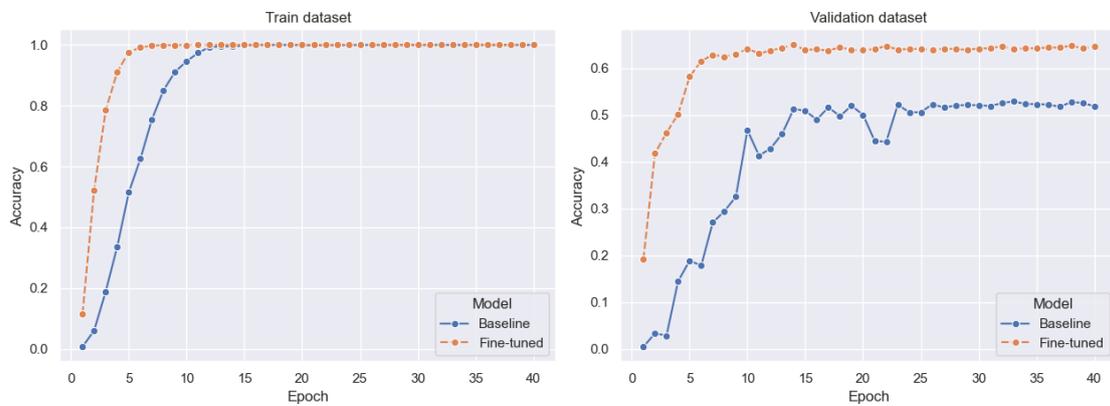

Figure 3: Comparison of the training processes with respect to the medium target dataset. The fine-tuned model's accuracy value converged at about `0.64` on the validation dataset, whereas the corresponding value for the baseline model was only about `0.52`.

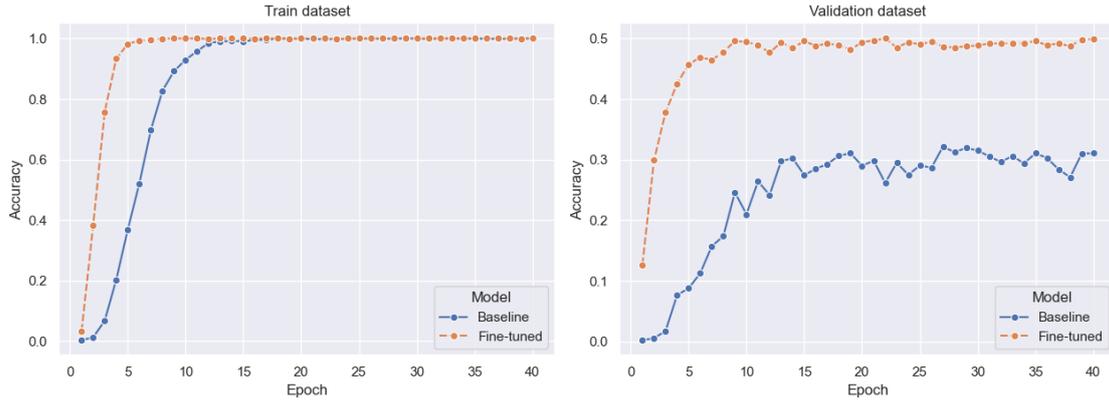

Figure 4: Comparison of the training processes with respect to the small target dataset. The fine-tuned model's accuracy value converged at about `0.49` on the validation dataset, whereas the corresponding value for the baseline model was only about `0.30`.

## 4.2 Comparison of the test dataset performances between the baseline and fine-tuned models

The fine-tuned models outperformed the corresponding baseline models substantially for both the loss and accuracy performances for each size of the training dataset:

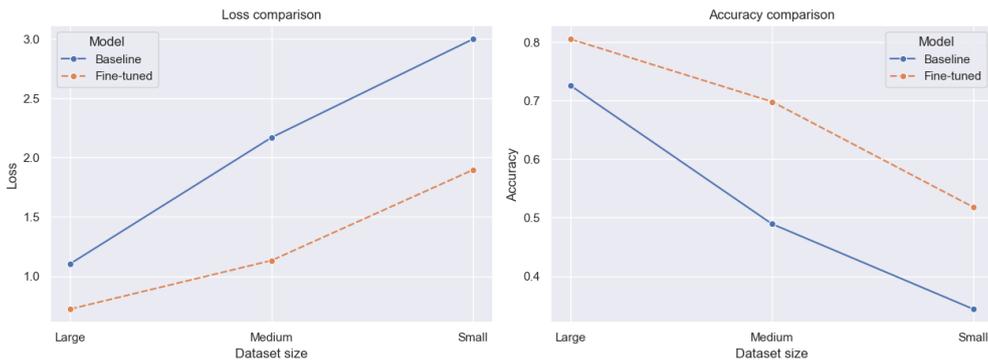

Figure 5: Comparison of the test performance scores between the baseline and fine-tuned models. As an example, in the case of the fine-tuned model trained on the small dataset, the loss was reduced from `2.9990` to `1.8959`, which corresponds to an improvement of about 63%.

The next figure illustrates the tendency of the improvements for the different training dataset sizes:

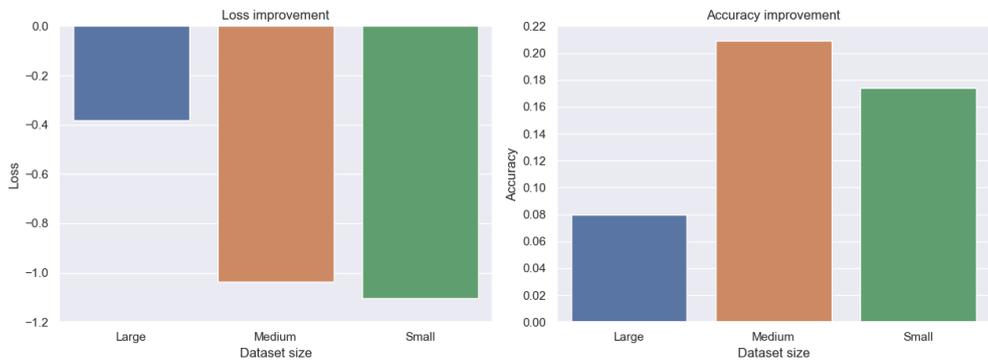

Figure 6: Improvements in the test performance scores. Accuracy of the fine-tuned models increased by `0.0795`, `0.2089` and `0.1742` for the large, medium and small datasets, respectively.

As expected, the improvements on the loss became greater, the smaller the size of the involved training dataset was. This tendency can be explained by the fact that a model trained on a smaller dataset possesses less generalization capability, and therefore, transfer learning has more effect. Concerning the accuracy metric, applying transfer learning had the most impact where the medium-sized dataset was used for training. That the greatest improvement was not achieved when the small training dataset was involved, does not need to raise major concerns as the improvement in the accuracy was only slightly smaller for the small dataset. Incidentally, the validation dataset exhibited the expected tendency that the improvements in the accuracy steadily grew with decreasing training dataset size: the accuracy performance increased from `0.0772` through `0.1316` to `0.1739` for the large, medium and small training dataset, respectively.

## 5 Conclusions

This work gives clear empirical evidence that transfer learning can be effectively applied to isolated SLR, i.e. an SLR model that utilizes transfer learning exhibits increased generalization capability. Specifically, it was demonstrated that knowledge from the American Sign Language domain could be successfully transferred to the German Sign Language domain. The predictive performances on the test dataset of the models involving transfer learning increased substantially as compared to the corresponding baseline models for each of the large, medium and small target training dataset: performance gains of 8%, 21% and 17% were achieved, respectively.

The effective application of transfer learning could not only be observed by higher final accuracy performances, but also by overall improved training behaviours. In all experiments conducted, the models that applied transfer learning showed the following additional advancements over the corresponding baseline models: 1) faster convergence meaning that fewer training epochs were required to reach the final accuracy performance, and 2) better initial accuracy performance on the validation dataset implying higher generalization capability from the outset.

Importantly, the beneficial effects of transfer learning arose from the fact that the dataset used for pre-training was an *SL* dataset and not just arbitrary video material. It is hoped that the insights gained will promote the development of SLR systems in particular for sign languages lacking adequate training data.

Computer-based sign language recognition is a challenging problem that can be tackled from various angles. Consequently, there are many opportunities for further research, such as studying the application of transfer learning to continuous SLR.

**Comments**

This paper arose out of my MSc thesis "Application of transfer learning to sign language recognition using an inflated 3D deep convolutional neural network" at The Open University submitted on 6[th] September 2020.

# References

Al-Shamayleh, A., Ahmad, R., Abushariah, M., Alam, K. and Jomhari, N. (2018) 'A systematic literature review on vision based gesture recognition techniques', *Multimedia Tools and Applications*, vol. 77, no. 21, pp. 28121-28184.

Athitsos, V., Neidle, C., Sclaroff, S., Nash, J., Stefan, A., Quan, Y. and Thangali, A. (2008) 'The American Sign Language Lexicon Video Dataset', *IEEE Computer Society Conference on Computer Vision and Pattern Recognition Workshops*. Anchorage, AK, 23-28 June. Washington, DC, IEEE Computer Society, pp. 1-8.

Bengio, Y., Courville, A. and Vincent, P. (2013) 'Representation Learning: A Review and New Perspectives', *IEEE Transactions on Pattern Analysis and Machine Intelligence*, vol. 35, no. 8, pp. 1798-1828.

Brooks, R. (2018) 'A Guide to the Different Types of Sign Language Around the World', *The Language Blog*, 10 May [Blog]. Available at https://k-international.com/blog/different-types-of-sign-language-around-the-world (Accessed 29 June 2020).

Cao, Z., Martinez, G., Simon, T., Wei, S. and Sheikh, Y. (2019) 'OpenPose: Realtime Multi-Person 2D Pose Estimation using Part Affinity Fields', *IEEE Transactions on Pattern Analysis and Machine Intelligence*, Early Access [Online]. DOI: 10.1109/TPAMI.2019.2929257 (Accessed 30 June 2020).

Carreira, J. and Zisserman, A. (2017) 'Quo Vadis, Action Recognition? A New Model and the Kinetics Dataset', *IEEE Conference on Computer Vision and Pattern Recognition*. Honolulu, 21-26 July. Washington, DC, IEEE Computer Society, pp. 4724-4733.

Cheok, M., Omar, J. and Jaward, Z. (2019) 'A review of hand gesture and sign language recognition techniques', *International Journal of Machine Learning and Cybernetics*, vol. 10, no. 1, pp. 131-153.

Er-Rady, A., Faizi, R., Thami, R. O. H. and Housni, H. (2017) 'Automatic sign language recognition: A survey', *International Conference on Advanced Technologies for Signal and Image Processing*. Fez, 22-24 May. Washington, DC, IEEE Computer Society, pp. 1-7.

Fang, B., Co, J. and Zhang, M. (2017) 'DeepASL: Enabling Ubiquitous and Non-Intrusive Word and Sentence-Level Sign Language Translation', *Proceedings of the 15th ACM Conference on Embedded Network Sensor Systems*. Delft, 6-8 November. New York, NY, Association for Computing Machinery, pp. 1-13.

Ibrahim, N. B., Selim, M. M. and Zayed, H. H. (2018) 'An Automatic Arabic Sign Language Recognition System (ArSLRS)', *Journal of King Saud University: Computer and Information Sciences*, vol. 30, no. 4, pp. 470-477.

Joze, H. and Koller, O. (2019) 'MS-ASL: A Large-Scale Data Set and Benchmark for Understanding American Sign Language', *British Machine Vision Conference*. Cardiff, 9-12 September. BMVA Press, pp. 1-16.

Kay, W., Carreira, J., Simonyan, K., Zhang, B., Hillier, C., Vijayanarasimhan, S., Viola, F., Green, T., Back, T., Natsev, P., Suleyman, M. and Zisserman, A. (2017) *The Kinetics Human Action Video Dataset* [Online]. arXiv:1705.06950 [cs.CV] (Accessed 1 September 2020).

Lapiak, J. A. (2013) *Rules of dominant, passive, and symmetrical hands* [Online]. Available at https://www.handspeak.com/learn/index.php?id=98 (Accessed 23 July 2020).

Starner, T. and Pentland, A. (1995) 'Real-time American Sign Language recognition from video using hidden Markov models', *Proceedings of International Symposium on Computer Vision*. Coral Gables, 21-23 November. Washington, DC, IEEE Computer Society, pp. 265-270.

Stokoe, W. C. (1960) *Sign language structure: an outline of the visual communication systems of the American deaf*, Buffalo, Dept. of Anthropology and Linguistics, University of Buffalo.

Szegedy, C., Liu, W., Jia, Y., Sermanet, P., Reed, S., Anguelov, D., Erhan, D., Vanhoucke, V. and Rabinovich, A. (2015) 'Going deeper with convolutions', *IEEE Conference on Computer Vision and Pattern Recognition*. Boston, 7-12 June. Washington, DC, IEEE Computer Society, pp. 1-9.

Szegedy, C., Vanhoucke, V., Ioffe, S., Shlens, J. and Wojna, Z. (2016) 'Rethinking the Inception Architecture for Computer Vision', *IEEE Conference on Computer Vision and Pattern Recognition*. Las Vegas, 27-30 June. Washington, DC, IEEE Computer Society, pp. 2818-2826.

Toengi (2020) Transfer learning for sign language recognition [Online]. Available at https://github.com/rtoengi/transfer-learning-for-sign-language-recognition (Accessed 4 September 2020).

Von Agris, U. and Kraiss, K. F. (2007) 'Towards a video corpus for signer-independent continuous sign language recognition', *Proceedings of GW 2007, 7th International Workshop on Gesture in Human-Computer Interaction and Simulation*, Lisbon, 23-25 May. pp. 1-6.

Woll, B. (2013) 'The History of Sign Language Linguistics', in Allan, K. *The Oxford Handbook of the History of Linguistics*, Oxford, Oxford University Press, pp. 91-104.

Xue, Q., Li, X., Wang, D. and Zhang, W. (2019) 'Deep forest-based monocular visual sign language recognition', *Applied Sciences*, vol. 9, no. 9, pp. 1-14.

Yosinski, J., Clune, J., Bengio, Y. and Lipson, H. (2014) 'How transferable are features in deep neural networks?', *Advances in Neural Information Processing Systems*, vol. 27, pp. 3320-3328.